\documentclass[runningheads]{llncs}

\usepackage{eccv}

\usepackage{eccvabbrv}

\usepackage{graphicx}
\usepackage{booktabs}
\usepackage{multirow}
\usepackage{makecell}
\usepackage{floatrow}
\newfloatcommand{capbtabbox}{table}[][\FBwidth]
\usepackage{subcaption}
\usepackage[accsupp]{axessibility}  

\usepackage{hyperref}

\usepackage{orcidlink}

\begin{document}

\title{HandDAGT: A Denoising Adaptive Graph Transformer for 3D Hand Pose Estimation} 
\titlerunning{A Denoising Adaptive Graph Transformer for 3D Hand Pose Estimation}
%
\author{Wencan Cheng\inst{1}\orcidID{0000-0002-7996-0236} \and Eunji Kim \inst{1} \orcidID{0009-0005-6039-1853} \and
Jong Hwan Ko\inst{2}\orcidID{0000-0003-4434-4318}}
\authorrunning{W. Cheng, E. Kim, J. H. Ko.}
%
\institute{Department of Artificial Intelligence, Sungkyunkwan University, South Korea \and
College of Information and Communication Engineering, Sungkyunkwan University, South Korea\\
\email{\{cwc1260,dmswldpvm,jhko\}@skku.edu}}
\maketitle

\begin{abstract}
The extraction of keypoint positions from input hand frames, known as 3D hand pose estimation, is crucial for various human-computer interaction applications. However, current approaches often struggle with the dynamic nature of self-occlusion of hands and intra-occlusion with interacting objects. To address this challenge, this paper proposes the Denoising Adaptive Graph Transformer, HandDAGT, for hand pose estimation.
The proposed HandDAGT leverages a transformer structure to thoroughly explore effective geometric features from input patches. Additionally, it incorporates a novel attention mechanism to adaptively weigh the contribution of kinematic correspondence and local geometric features for the estimation of specific keypoints. This attribute enables the model to adaptively employ kinematic and local information based on the occlusion situation, enhancing its robustness and accuracy.
Furthermore, we introduce a novel denoising training strategy aimed at improving the model's robust performance in the face of occlusion challenges. 
Experimental results show that the proposed model significantly outperforms the existing methods on four challenging hand pose benchmark datasets. Codes and pre-trained models are publicly available at \url{https://github.com/cwc1260/HandDAGT}.
\end{abstract}

\section{Introduction}
\label{sec:intro}
\begin{figure}[!h]
\centering
\includegraphics[width=0.8\linewidth]{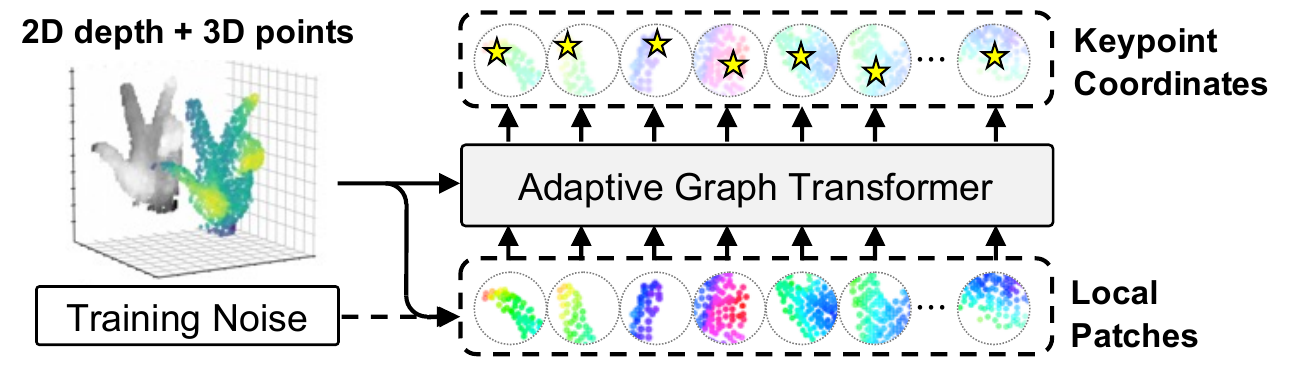}
\caption{Illustration of the HandDAGT concept. The model feeds 3D local patches cropped from input depth images and corresponding point clouds to the adaptive graph transformer for the keypoint coordinate estimation. The local patches are disturbed by the random Gaussian noise during training for a robust performance.}
\label{fig:concept}
\end{figure}
3D hand pose estimation has emerged as a pivotal element in various human-computer interaction systems, providing essential keypoints for human hands. Accurate localization of hand keypoints enables precise understanding and effective analysis of human motion and behavior. With the advancements in deep learning techniques and the proliferation of low-cost depth cameras, significant progress has been made in depth image-based 3D hand pose estimation.

Early deep learning-based methods showcased the efficacy of conventional convolutional neural networks (CNNs) in hand pose estimation tasks \cite{tompson2014real, ge2016robust, guo2017region, ren2019srn, chen2020pose, fang2020jgr, ren2021pose, du2019crossinfonet}. Subsequently, CNN architectures leveraging different modalities, such as point clouds \cite{ge2018hand, ge2018point, cheng2021handfoldingnet, cheng2023handr2n2}, voxelized representations \cite{ge20173d, moon2018v2v}, or multi-modal combination \cite{ren2023two}, have been extensively explored. Beyond CNN, the transformer-based approaches \cite{huang2020hand, wang2023handgcnformer} also showed the feasibility in the hand pose estimation field. Despite the notable successes achieved by these approaches, achieving accurate estimation remains challenging in scenarios such as self-occlusion and hand-object occlusion. Recent studies \cite{cheng2023handr2n2, ren2023two, wang2023handgcnformer}, have endeavored to introduce a static graph structure to capture kinematic correspondence between keypoints, enabling the reconstruction of occluded keypoints using corresponding keypoints' information. However, static graph structures fall short of capturing the dynamics of various occlusions.

To address these challenges, we propose a novel denoising adaptive graph transformer, named HandDAGT, to adaptively weigh local detail and kinematic correspondence for accurate hand pose estimation across various occlusion conditions. The proposed transformer can learn precise local geometric information of the hand as the traditional local transformers \cite{zhao2021point, gao2022lft, wu2021centroid}. Importantly, the transformer is meticulously designed to calculate the geometric feature and kinematic topology information directly in the 3D keypoint space. This design is made to circumvent 2D-to-3D non-linearity that arises when operating in 2D space \cite{cheng2023handr2n2}. Moreover, for accurate estimation, it integrates a novel adaptive attention module that adaptively activates local geometric information or information from kinematically corresponding keypoints depending on whether keypoints are well-presented or occluded. Furthermore, to enhance the robustness and accuracy of estimation under occlusion, we introduce a novel denoising training strategy as it compels the model to correct estimations even with disturbed patches.

We evaluate HandDAGT on four challenging benchmarks, including single-hand ICVL \cite{tang2014latent} and NYU \cite{tompson2014real} dataset, and hand-object DexYCB \cite{chao2021dexycb} and HO3D \cite{hampali2020honnotate} datasets. The results show that our network achieves the state-of-the-art performance with the lowest mean distance errors of 5.66 mm and 7.12 mm on the single-hand ICVL and NYU datasets, respectively. The model also achieves state-of-the-art performance on the hand-object DexYCB and HO3D datasets with the lowest error of 8.03 mm and 1.81 cm, respectively.

Our main contributions are summarised as follows:
\begin{itemize}
\item We propose a novel transformer-based architecture, HandDAGT, for accurate 3D hand pose estimation that utilizes the depth image and point cloud as a multi-modal input. 
\item We propose a novel adaptive attention module designed to weigh the local geometric information and kinematically evolved keypoint feature for high reliability to occlusions.
\item We introduce a novel denoising training strategy for hand pose estimation to improve robustness and accuracy.
\item We perform comprehensive experiments on big and challenging benchmarks that present the new state-of-the-art performance of our proposed method.
\end{itemize}

\section{Related Work}
\subsection{3D Hand Pose Estimation Based on Depth Image.}

In the realm of 3D hand pose estimation utilizing depth images, traditional 2D CNN-based approaches \cite{tompson2014real, ge2016robust, guo2017region, ren2019srn, chen2020pose, fang2020jgr, ren2021pose, du2019crossinfonet} have garnered widespread adoption due to their straightforward implementation. However, they are encumbered by inherent limitations such as the inability to fully discern the intricacies of the 3D structure and their reliance on the camera's perspective.

To surmount these constraints, 3D CNN-based methodologies \cite{ge20173d, moon2018v2v} were introduced, employing 3D voxelized representations of depth images to encapsulate volumetric data. Despite their advancements in 3D hand pose estimation, these techniques demand substantial memory and computational resources, thus curtailing their practical applicability.

In contrast, PointNet-based strategies \cite{ge2018hand, ge2018point, chen2018shpr, li2019point, cheng2021handfoldingnet} analyze point clouds, offering a precise depiction of the 3D structure. PointNet, a deep learning framework adept at handling irregular and unstructured point clouds, was initially employed for hand pose estimation in HandPointNet \cite{ge2018hand}. Subsequent enhancements include the Point-to-Point model \cite{ge2018point} and SHPR-Net \cite{chen2018shpr}, which refine performance by generating point-wise probability distributions. Notably, SHPR-Net \cite{chen2018shpr} amalgamated HandPointNet with an auxiliary semantic segmentation subnetwork to augment performance. Recently, HandFoldingNet \cite{cheng2021handfoldingnet} introduced a folding concept to reshape a pre-defined 2D hand skeleton into hand poses, thereby enhancing estimation accuracy. Subsequently, HandR2N2 \cite{cheng2023handr2n2} proposed a folding-based recurrent structure to iteratively optimize the keypoint positions by searching local point regions. However, a notable drawback of point clouds is the computational overhead of querying neighbors from a dense point set for convolution. Therefore, the existing methods commonly use a sparse point cloud, which restricts the performance. 

In order to bridge the shortcomings of images and point clouds, IPNet \cite{ren2023two} suggests the utilization of multi-modal representations, combining 2D depth images and 3D point clouds. By doing so, the model adeptly extracts dense detail information while effectively capturing 3D spatial features, thereby facilitating accurate 3D hand pose estimation. 

Essentially, the above-mentioned methods are all CNN-induced methods but with different input modalities. In contrast, our method introduces a novel adaptive graph transformer architecture for precise local information extraction and dynamic kinematics with the power of the transformer.

\subsection{Transformers in Hand Pose Estimation}
With the development of the attention mechanism and the introduction of the transformer model \cite{vaswani2017attention}, various domains within image processing have increasingly embraced its application, such as image classification \cite{dosovitskiy2020image, liu2021swin} and object detection \cite{carion2020end, zhu2020deformable}. Given the transformer model's ability to capture non-local features, which proves beneficial for hand pose estimation, numerous studies \cite{lin2021end, huang2020hand, lin2021mesh, wang2023handgcnformer, hampali2022keypoint, li2022interacting} have endeavored to integrate it into hand pose estimation domain. As a closely related work to ours, Hand-Transformer \cite{huang2020hand} introduces a non-autoregressive transformer to concurrently estimate keypoint positions. Nevertheless, this approach of independent keypoint estimation overlooks the intrinsic kinematics shared among keypoints. Consequently, HandGCNFormer \cite{wang2023handgcnformer} integrates graph convolutional networks (GCNs) as its core to explicitly model the kinematics among keypoints, thereby enhancing performance. However, its static graph is not able to capture dynamic kinematic correspondence between keypoints under complex occlusion scenarios. In contrast, the proposed HandDAGT is able to dynamically weigh between kinematic correspondence and local detail according to the occlusion situation.

\begin{figure}[!t]
\centering
\includegraphics[width=0.8\linewidth]{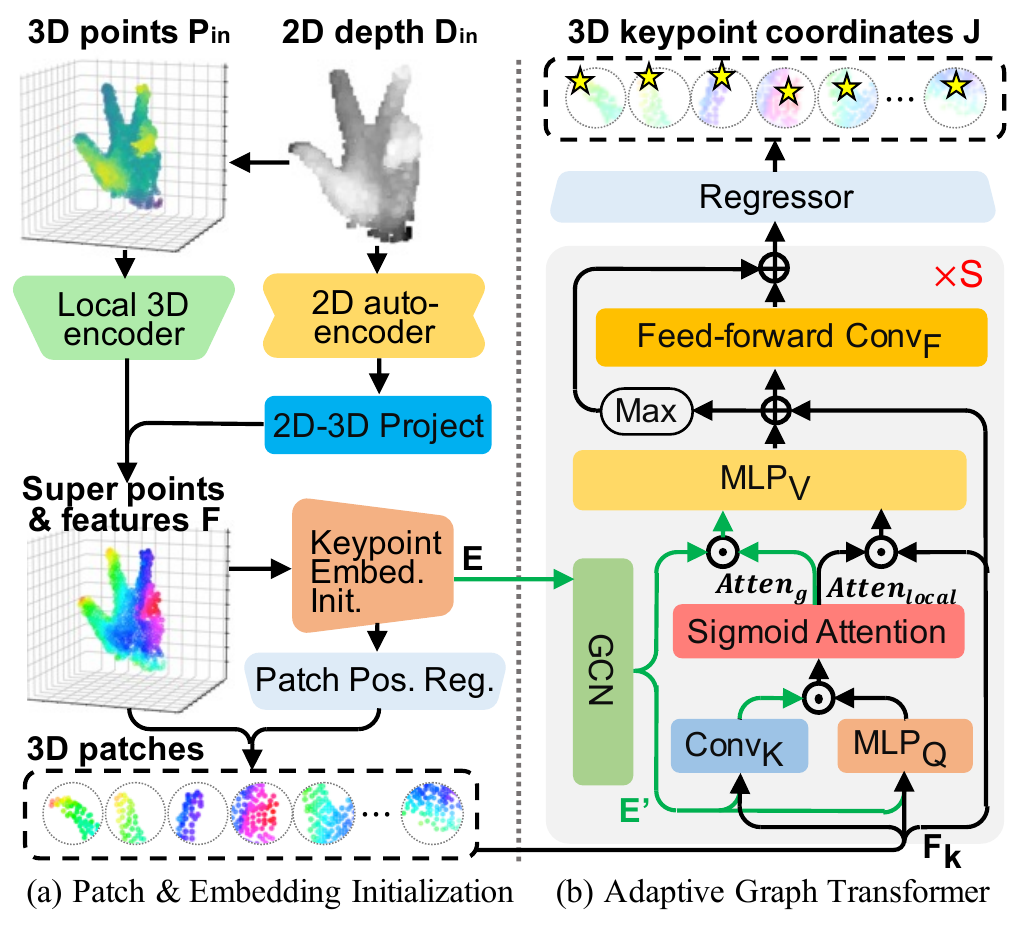}
\caption{The HandDAGT architecture. HandDAGT takes a 2D depth image and the sub-sampled point cloud as the input. The PointNet-based local 3D encoder and the 2D auto-encoder extracts local 3D features and local 2D features, respectively. Then, the 2D features are projected into 3D space to fuse with 3D features forming the super point features F. Based on the super points, keypoint embeddings E and 3D patches are extracted as input to the novel adaptive graph transformer to estimate accurate 3D keypoint coordinates by leveraging the dynamic kinematic correspondences and local details. Notably, during the training stage, the local 3D patches are shifted by random noises in order to enforce the model providing robust estimations.}
\label{fig:architecture}
\end{figure}

\section{Method}
The HandDAGT architecture is presented in Figure~\ref{fig:architecture}. HandDAGT first samples a super point set with fused features from both image and point cloud, thus it can generate strong keypoint embeddings and informative local patches. The embeddings and patches are further fed as queries and keys to our key component, the adaptive graph transformer. The transformer implements a novel attention mechanism that adaptively weighs queries and keys for the accurate keypoint coordinate estimation.

\subsection{Super Points Generation}
Following the previous work IPNet \cite{ren2023two}, HandDAGT accepts the multimodal inputs, which are a hand depth image $\mathbf{D}_{in} \in \mathbb{R}^{H \times W}$ and a set of randomly sampled 3D point coordinates $\mathbf{P}_{in} \in \mathbb{R}^{N \times 3}$. The outputs from HandDAGT are 3D keypoint coordinates $\mathbf{J} \in \mathbb{R}^{J \times 3}$. Since the input point set is irregular and orderless, we exploit the point set convolutional layer \cite{qi2017pointnet++, liu2019flownet3d} as a local 3D encoder to extract a subsampled 3D super point set $\mathbf{P'} \in \mathbb{R}^{N/2 \times 3}$ with corresponding 3D local geometric features $\mathbf{F}_{3d} \in \mathbb{R}^{N/2 \times d_{3d}}$. The depth image is supplied into a ConvNeXt-based autoencoder to generate a 2D local visual feature map $\mathbf{F}_{2d} \in \mathbb{R}^{H/2 \times W/2 \times d_{2d}}$. Subsequently, the 2D local visual features are projected as 3D point features $\mathbf{F}_{2d-3d} \in \mathbb{R}^{N/2 \times d_{2d}}$ through the 2D-3D projection technique proposed by IPNet \cite{ren2023two}. Afterward, the projected features are concatenated to the super points as $\mathbf{F} \in \mathbb{R}^{N/2 \times (d_{2d}+d_{3d})}$. Then the super points are further fed into the initialization module to generate initial keypoint embeddings and patch positions for the transformer. 

\subsection{Initialization of Keypoint Embedding and Patch Position}

In our HandDAGT framework, the keypoint embeddings play a critical role in capturing the kinematic topology between keypoints. To achieve this, we draw upon the concept of joint-wise embeddings initially proposed by HandFoldingNet \cite{cheng2021handfoldingnet} to obtain the initial keypoint embeddings.

The keypoint embedding initialization module begins by feeding the super points with features $\mathbf{F}$ into two point set convolutional layers to generate 3D global vectors. The initialization of keypoint embeddings is then sequentially conducted through a three-layer bias-induced layer (BIL) \cite{cheng2019point}. 
The  global vector is replicated $J$ times and supplied to the BILs. This operation yields embedding for the $J$ individual keypoints. Notably, the BIL furnishes keypoint-wise independent biases, akin to learnable positional embeddings. 

After the acquisition of the keypoint embeddings, a linear transformation is applied to project the high-dimensional embeddings into the 3D space as the initial positions for collecting patches. The 3D point patches are acquired by grouping $K$-nearest super points with their corresponding features around each patch position. 

\subsection{Adaptive Graph Transformer Decoder}
The adaptive transformer decoder is the key component in our proposed framework as it queries the local information for accurate keypoint estimation with the assist of dynamic kinematic topology based on the keypoint embeddings, as illustrated in Figure \ref{fig:architecture}. 

The keypoint embeddings are first augmented through a GCN, forming the evolved embeddings,
\begin{equation}
\mathbf{E'} = ReLU(\mathbf{A}\mathbf{E}\mathbf{W}),
\end{equation}
where $\mathbf{W} \in \mathbb{R}^{d_{hid} \times d_{hid}}$ is the trainable transformation weights and $\mathbf{A} \in \mathbb{R}^{J \times J \times d_{hid}}$ is the learned kinematic correspondence matrix that models connection strength between keypoints in each channel. 

Subsequently, each super point in each patch and its corresponding feature are concatenated with the evolved keypoint embedding, forming the input to the novel attention mechanism of the proposed transformer,
\begin{equation}
\mathbf{Q}_{k} = \mathop{MLP_Q}([\mathbf{E'}, \mathbf{F}_k]),
\label{eq.q}
\end{equation}
\begin{equation}
\mathbf{K} = \mathop{set\_conv_K}\limits_{1 \leq k \leq K} ([\mathbf{E'},\mathbf{F}_k]), 
\label{eq.k}
\end{equation}
\begin{equation}
[\mathbf{Att^{local}_k}, \mathbf{Att^{g}_k}] = \sigma(\mathbf{Q}_{k} \odot \mathbf{K}),
\label{eq.score}
\end{equation}
\begin{equation}
\mathbf{V}_{k} =\mathop{MLP_V} ([\mathbf{E'} \odot \mathbf{Att^{g}_k},\mathbf{F}_k \odot \mathbf{Att^{local}_k}]),
\end{equation}
where $\mathop{MLP}$ is the multi-layer perceptron, $\mathop{set\_conv}$ is the point set convolutional layer, $\sigma$ is the sigmoid activation function, `$[\cdot, \cdot]$' is the concatenation operation, and $\odot$ is the Hadamard product. We notice that the semantic difference between local points and keypoint embeddings affects the effectiveness of attention, we enhance the queries (local points) \textbf{Q} with the evolved keypoint embedding while the keys (keypoint embeddings) \textbf{K} are fused with the local observation. Furthermore, unlike conventional attention that commonly utilizes dot product with softmax, our attention score is obtained by the Hadamard product with a sigmoid as the activation as shown in Equation \ref{eq.score}. Moreover, due to the channel-wise multiplication of the Hadamard product, the proposed attention mechanism can be regarded as inherently employing the multi-head attention with the same number of attention heads as the channel depth. It is worth mentioning that the attention produces two individual scores for the local points and graph-evolved key embeddings. With these two scores, the proposed transformer is able to adaptively focus more on kinematic topology or local information of the keypoints depending on whether the keypoints are occluded or well-represented by the local points.

Following the standard attention block, a residual mechanism is introduced to accumulate the computed value \textbf{V} and the input local points. Afterward, another point set convolution is applied as the feed-forward module to aggregate the computed value \textbf{V} to an updated key embedding for position regression of each keypoint,
\begin{equation}
\mathbf{\widetilde{E}} = \mathop{set\_conv_F}\limits_{1 \leq k \leq K} (\mathbf{V_k}+\mathbf{F_k})+\mathop{MAX_F}\limits_{1 \leq k \leq K} (\mathbf{V_k}+\mathbf{F_k}),
\end{equation}
where $\mathop{MAX_F}$ indicates the max-pooling operation for pursuing the residual accumulation. At the end, a trainable linear transformation is followed to project the updated keypoint embeddings to the 3D keypoint coordinates. 

For an improved estimation performance, the proposed transformer is stacked multiple times to progressively optimize the keypoint position. The regressed positions from the previous transformers are used for grouping new patches as the input to the subsequent transformer stacks. The optimal number of stacked transformers will be discussed in Section \ref{sec:abla}.

\subsection{Denosing Training and Inference}
To improve the accuracy and robustness of the model against noise and occlusion, we propose a novel denoising training strategy. The initial patch positions are firstly disturbed by the introduced random noise. The noise variances during training are controlled by the same linear scheduler as in recent diffusion models \cite{ho2020denoising, chen2022diffusiondet}.
However, in contrast to the diffusion models that introduce noise in both the training and inference phases, our method only introduces noise in the training stage. Thus, the training criterion in our method is formally defined as:
\begin{equation}
\mathcal L = \sum L1_{smooth}(\textbf{D}_{T_1}(\textbf{J}_0 + \mathcal N) - \textbf{J}^*) + \sum_2^S \sum L1_{smooth}(\textbf{D}_{T_s}(\textbf{J}_{s-1}) - \textbf{J}^*),
\end{equation}
where $\mathcal N$ is the injected noise, $S$ denotes the number of stacked denosing transformers $\textbf{D}_{T_s}$, and $\textbf{J}_{s}$ is the output from the s-th denosing transformer $\textbf{D}_{T_s}(\textbf{J}_{s-1})$. Note that, the noise is only injected in the input of the first denosing transformer.
Following the previous works \cite{ren2019srn, cheng2021handfoldingnet}, we adopt the smooth L1 loss because of its less sensitivity to the outliers. The smooth L1 loss is defined as:
\begin{equation}
	L1_{smooth}(\textbf{x}) = \begin{cases}
	50\textbf{x}^2, &|\textbf{x}|<0.01\\
	|\textbf{x}|-0.005, &otherwise
		   \end{cases}.
\end{equation}

\section{Experiments}
\subsection{Experiment Settings}
In our experiments, we utilized an NVIDIA A100 GPU with PyTorch for training and evaluating the model. We employed the AdamW optimizer \cite{loshchilov2017decoupled} with parameters set to beta$_1$ = 0.5, beta$_2$ = 0.999, and a learning rate $\alpha$ = 0.001.
The input images were resized to 128, and the number of input points to the network was sampled to 1,024. We set the batch size to 32. To mitigate overfitting, we employed online data augmentation techniques, including random rotation (in the range of [-180.0, 180.0] degrees), 3D scaling (within the range of [0.9, 1.1]), and 3D translation (within the range of [-10, 10] mm).
For training on the hand-object datasets, we conducted training for 30 epochs with a learning rate decay of 0.1 after every 10 epochs. In the case of training on the single-hand dataset, we extended the training duration to 50 epochs with a similar learning rate decay strategy, applying the decay after 30 epochs. 

\subsection{Datasets and Evaluation Metrics}
\noindent
{\bf Single-hand Dataset.} {\bf ICVL} \cite{tang2014latent} comprises 22,000 training depth frames and 1,600 testing depth frames. Each frame includes annotations for $J = 16$ keypoints, including one keypoint for the palm and three keypoints for each finger. 
{\bf NYU} \cite{tompson2014real} provides depth images of single hands captured from three distinct viewpoints using the PrimeSense 3D sensor. Each viewpoint comprises 72,000 frames for training and 8,000 frames for testing. In line with recent studies \cite{cheng2021handfoldingnet, ge2018point, ge2018hand}, a single view is utilized for both training and testing purposes. Furthermore, for evaluation, 14 keypoints out of a total of 36 annotated keypoints are selected, maintaining consistency with prior works.

\noindent
{\bf Hand-object Dataset.} {\bf DexYCB} \cite{chao2021dexycb} is a recently released hand-object dataset, comprising 582,000 image frames annotated with 21 keypoints. It incorporates data from 10 distinct subjects interacting with 20 YCB objects across 8 camera views. Notably, the dataset defined four official split protocols: S0 - seen subjects, camera views, grasped objects; S1 - unseen subjects; S2 - unseen camera views; S3 - unseen grasped objects. 
{\bf HO3D} \cite{hampali2020honnotate} is another challenging dataset renowned for its precise hand-object pose during interactions. The widely utilized version, HO3D\_v2, comprises 66,034 training images and 11,524 testing images gathered from 10 subjects engaging with 10 distinct objects. Evaluation for HO3D\_v2 is typically conducted through an online submission platform dedicated to this dataset.

\noindent
{\bf Evaluation metrics.}
We employ two commonly used metrics, the mean keypoint error, and the success rate, to evaluate the performance of hand pose estimation. The mean keypoint error quantifies the average Euclidean distance between the estimated keypoint positions and ground-truth ones for all the keypoint over the entire testing set. The success rate indicates the percentage of good frames with a mean keypoint error of less than a given distance threshold.

\subsection{Comparison with State-of-the-Art Methods}
\begin{table}[t!] \small
\caption{Comparison of the proposed method with previous state-of-the-art methods on the single-hand ICVL and NYU datasets. Input indicates the input type of 2D depth image (D), 3D voxels (V), or 3D point cloud (P).}
\centering
\begin{tabular}{c|m{2cm}<{\centering}m{2cm}<{\centering}|c}
\hline
 \multirow{2}{*}{Method} & \multicolumn{2}{c|}{Mean keypoint error (mm)}& \multirow{2}{*}{Input}\\
\cline{2-3}
                       & ICVL       & NYU       &   \\
\hline

Ren-9x6x6 \cite{guo2017region}       &7.31         & 12.69         &D  \\
DeepPrior++ \cite{oberweger2017deepprior++}& 8.1   & 12.24         &D  \\
Pose-Ren \cite{chen2020pose}         &6.79         & 11.81         &D  \\
DenseReg \cite{wan2018dense}         &7.3          & 10.2          &D  \\
CrossInfoNet \cite{du2019crossinfonet}&6.73        & 10.08         &D  \\
JGR-P2O \cite{fang2020jgr}           &6.02         & 8.29          &D  \\
SSRN \cite{ren2021spatial}           &6.01         & 7.37          &D  \\
PHG \cite{ren2021pose}               &5.97         & 7.39          &D  \\
\hline
HandPointNet \cite{ge2018hand}       &6.94         & 10.54         &P  \\
Hand-Transformer \cite{huang2020hand}   &6.47    & 9.80       &P  \\
Point-to-Point \cite{ge2018point}    &6.3          & 9.10          &P  \\
V2V \cite{moon2018v2v}               &6.28         & 8.42          &V  \\
HandFolding \cite{cheng2021handfoldingnet}   &5.95    & 8.58       &P  \\
HandR2N2 \cite{cheng2023handr2n2} & 5.70 & 7.27 & P \\
IPNet \cite{ren2023two}    &5.76& 7.17  &D+P  \\
\hline
HandDAGT (Ours)    &\textbf{5.66}& \textbf{7.12}  &D+P  \\
\hline
\end{tabular}
\label{tab:sota}
\end{table}
\begin{figure}[t!]
\centering
\includegraphics[width=\linewidth]{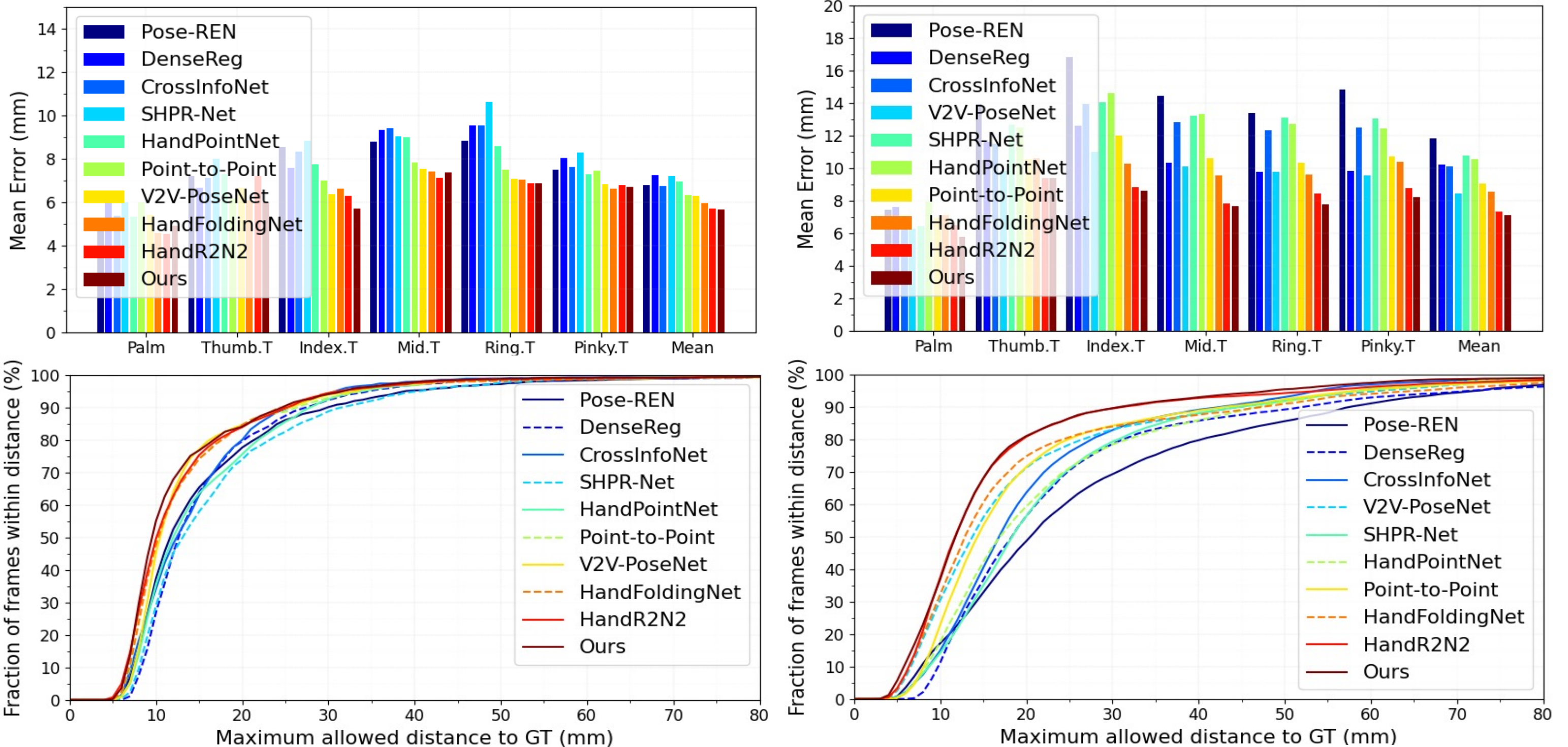}
\caption{Comparison with the state-of-the-art methods using the ICVL (left) and NYU (right) dataset. The per keypoint error (top) and success rate (bottom) are shown in this figure.}
\label{fig:threshold}
\end{figure}

\begin{figure}
\centering
\includegraphics[width=0.97\linewidth]{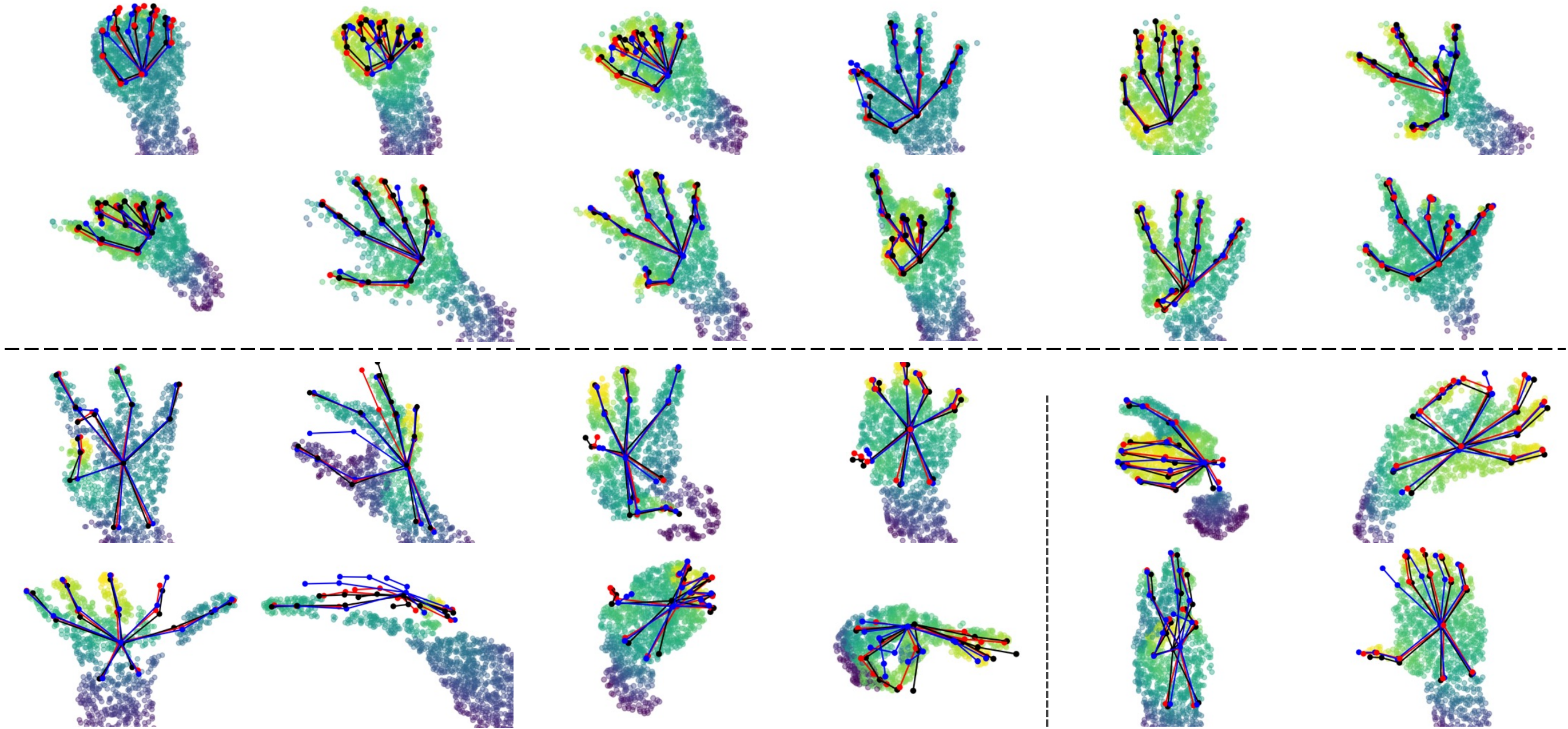}
\caption{Qualitative results of HandDAGT on the ICVL (top) and NYU (bottom) datasets. Hand-depth images are transformed into 3D points to clearly illustrate occlusions. Ground truth keypoints are represented in black, results from the comparative HandR2N2 \cite{cheng2023handr2n2} model are shown in blue, and the estimated keypoint coordinates of our model are depicted in red. The bottom figures showcase self-occluded/truncated cases (left) and well-performed cases without occlusion (right) in the NYU dataset.}
\label{fig:vis}
\end{figure}

\textbf{Single Hand.} We compare HandDAGT on the ICVL and NYU dataset with other state-of-the-art methods, including methods with 2D depth images as input: DeepPrior++ \cite{oberweger2017deepprior++},  Ren-9x6x6 \cite{guo2017region}, Pose-Ren \cite{chen2020pose}, DenseReg \cite{wan2018dense}, CrossInfoNet \cite{du2019crossinfonet}, JGR-P2O \cite{fang2020jgr}, spatial-aware stacked regression network (SSRN)  \cite{ren2021spatial} and pose-guided hierarchical graph network (PHG) \cite{ren2021pose}, methods with 3D point cloud or voxels as input: HandPointNet \cite{ge2018hand}, Point-to-Point \cite{ge2018point}, Hand-Transformer \cite{huang2020hand}, V2V \cite{moon2018v2v}, HandFoldingNet \cite{cheng2021handfoldingnet} and HandR2N2 \cite{cheng2023handr2n2}, and method with multi-modal input, IPNet \cite{ren2023two}.

Table \ref{tab:sota}  summarizes the performance in terms of mean keypoint error on two single-hand datasets. The results highlight that HandDAGT achieves a new state-of-the-art record with the lowest mean distance errors of 5.66 mm and 7.12 mm on the challenging ICVL and NYU datasets, respectively.
Furthermore, the outcomes demonstrate that HandDAGT significantly outperforms other 2D image-based methods by substantial margins. This superiority is attributed to HandDAGT's direct processing in the 3D space, which circumvents the highly non-linear mapping problem associated with estimating from 2D images. The qualitative results illustrated in Figure \ref{fig:vis} also show the evidence of the HandDAGT's superiority in both occluded and non-occluded cases. Figure \ref{fig:threshold} represents that our method significantly outperforms other methods in terms of success rate when the error threshold is lower than 13 and 8 mm on the ICVL and NYU datasets, respectively. 

\begin{figure}
\centering
\includegraphics[width=\linewidth]{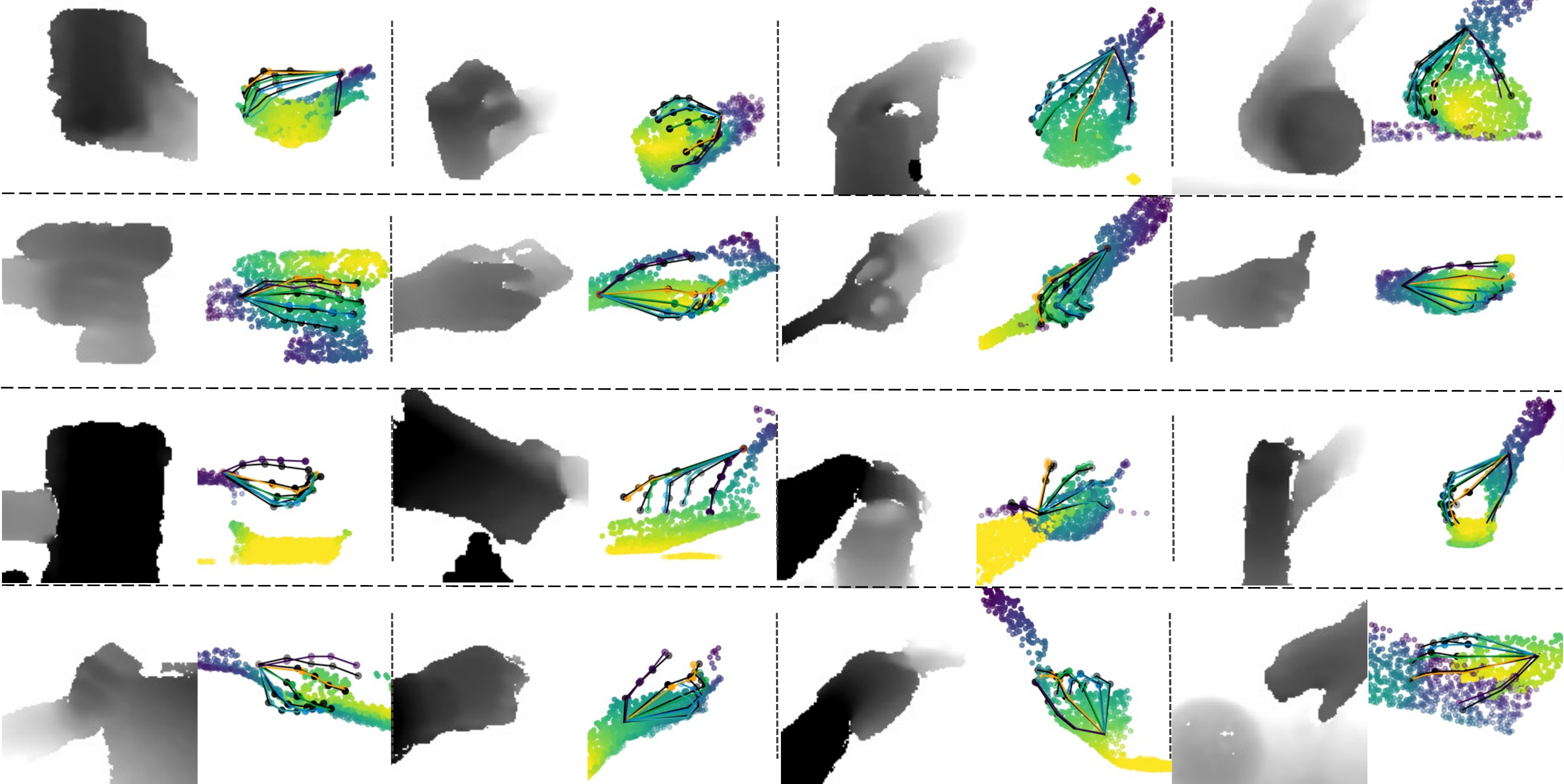}
\caption{Qualitative results of HandDAGT on the DexYCB dataset including different grabbing poses (top), interacting objects (2nd row), object occlusions (3rd row), and self-occlusions (bottom). Hand-depth images (left) are transformed into 3D points (right) in order to clearly present occlusions as shown in the figure. Ground truth is shown in black and the estimated keypoint coordinates of our model are shown in colors.}
\label{fig:ycb}
\end{figure}
\noindent
\textbf{Hand-Object.} We compare HandDAGT on the hand-object dataset DexYCB with other state-of-the-art method on the official dataset split protocols, including A2J \cite{xiong2019a2j}, Spurr et al. \cite{spurr2020weakly}, METRO \cite{lin2021end}, Tse et al. \cite{tse2022collaborative}, HandOccNet \cite{park2022handoccnet} and IPNet \cite{ren2023two}. As shown in Table \ref{tab:dexycb}, HandDAGT significantly outperforms previous SOTA methods on all four protocols. Qualitative results visualized in Figure \ref{fig:ycb} also reveal that HandDAGT can estimate accurate poses from hand-object interaction scenarios with various occlusions. We also compared HandDAGT on another the hand-object dataset HO3D with state-of-the-art method Hybrik \cite{li2021hybrik}, ArtiBoost \cite{yang2022artiboost}, HandOccNet \cite{park2022handoccnet}, HandVoxNet++ \cite{malik2021handvoxnet++} and IPNet \cite{ren2023two}. The comparison results are presented in Table \ref{tab:ho3d}. The results demonstrate that our HandDAGT can provide the comparable performance with the same mean distance error compared with the state-of-the-art model IPNet. 

\begin{table}[t!] \small
\caption{Comparison of the proposed method with previous state-of-the-art methods on the DexYCB datasets.}
\centering
\tabcolsep=3pt
\begin{tabular}{c|cccc|c|c}
\hline
 \multirow{2}{*}{Method} & \multicolumn{5}{c|}{Mean keypoint error (mm)}& \multirow{2}{*}{Input}\\
\cline{2-6}
            & S0    &S1     & S2    & S3    & AVG   &  \\
\hline
A2J \cite{xiong2019a2j}        & 23.93 & 25.57 & 27.65 & 24.92 & 25.52 & D\\
Spurr et al. \cite{spurr2020weakly} & 17.34 & 22.26 & 25.49 & 18.44 & 18.44 & RGB\\
METRO    \cite{lin2021end}   & 15.24 &-&-&-&-& RGB \\
Tse et al. \cite{tse2022collaborative} & 16.05 & 21.22 & 27.01 & 17.93 & 20.55 & RGB \\
HandOcc \cite{park2022handoccnet}  & 14.04 &-&-&-&-& RGB \\
IPNet \cite{ren2023two}      & 8.03  & 9.01  & 8.60  & 7.80  & 8.36  & D+P\\
\hline
HandDAGT (Ours)         & \textbf{7.72}  & \textbf{8.68}  & \textbf{8.22}  & \textbf{7.52} & \textbf{8.03}  & D+P \\
\hline
\end{tabular}
\label{tab:dexycb}
\end{table}

\begin{table}[t!] \small
\caption{Comparison of the proposed method with previous state-of-the-art methods on the HO3D datasets.}
\centering
\tabcolsep=3pt
\begin{tabular}{c|c|c}
\hline
Method & Mean keypoint error (mm)& Input\\
\hline
Hybrik \cite{li2021hybrik}      & 2.89 & RGB\\
ArtiBoost \cite{yang2022artiboost}   & 2.53 & RGB\\
HandOccNet   \cite{park2022handoccnet}   & 2.49 & RGB \\
HandVoxNet++ \cite{malik2021handvoxnet++}& 2.46 & V \\
IPNet \cite{ren2023two}      & 1.81  & D+P\\
\hline
HandDAGT (Ours)         & \textbf{1.81}  & D+P \\
\hline
\end{tabular}
\label{tab:ho3d}
\end{table}

\subsection{Ablation Study}
\label{sec:abla}
\noindent
\textbf{Effectiveness of the proposed attention mechanism.}
To assess the effectiveness and necessity of the proposed attention mechanism, we conducted a series of ablation experiments as follows:

1) No attention mechanism: Only the feed-forward module is retained for pose estimation.
2) Conventional attention mechanism: The conventional attention mechanism that uses dot product and softmax is implemented.
3) No sigmoid attention: Upon our attention mechanism, the sigmoid attention (Equation \ref{eq.score}) is replaced as the dot product and softmax as in the conventional attention.
4) No enhanced \textbf{Q}, \textbf{K}: Upon our attention mechanism, the enhancement for query (Equation \ref{eq.q}) and key (Equation \ref{eq.k}) are removed. 
5) Only sigmoid attention on local points $\mathbf{Att^{local}_k}$: Attention mechanism solely for local points is activated.
6) Only sigmoid attention on kinematic topology $\mathbf{Att^{g}_k}$: Attention mechanism concentrating exclusively on kinematic topology is activated.
7) Full implementation: Both sigmoid attentions on local points and kinematic topology are activated.
It's worth noting that the denoising training strategy is applied for all ablations, and the number of stacked transformers is set to 3 for consistency and comparison across experiments.


\begin{table}[t]
\caption{Ablations of different configurations of the transformer. All the ablation models are trained and tested on the NYU dataset. `C' denotes the conventional attention mechanism. `S' denotes the proposed sigmoid attention.}
\label{tab:abl}
\small
\centering
\begin{tabular}{c|cccc|c}
\hline
\multirow{3}{*}{ID} & \multicolumn{4}{c|}{Configurations} & \multirow{3}{*}{\makecell{Mean keypoint \\ error (mm)}} \\
\cline{2-5}
& Attention & Enhanced & \multirow{2}{*}{$\mathbf{Att^{local}_k}$}& \multirow{2}{*}{$\mathbf{Att^{g}_k}$}& \\
& type & \textbf{Q}, \textbf{K}  & & \\
\hline
1) & $\times$ & $\times$ & $\times$ & $\times$ &  7.43\\
\hline
2) & C & $\times$ & $\surd$ & $\surd$ & 10.66\\
3) & C & $\surd$ & $\surd$ & $\surd$ & 10.38\\
4) & S & $\times$ & $\surd$ & $\surd$ & 7.30\\
5) & S & $\surd$ & $\surd$ & $\times$ & 7.24 \\
6) & S & $\surd$ & $\times$ & $\surd$ & 7.21 \\
7) & S & $\surd$ & $\surd$ & $\surd$ & \textbf{7.12} \\
\hline
\end{tabular}
\end{table} 

\begin{figure}[t]
\centering
\includegraphics[width=\linewidth]{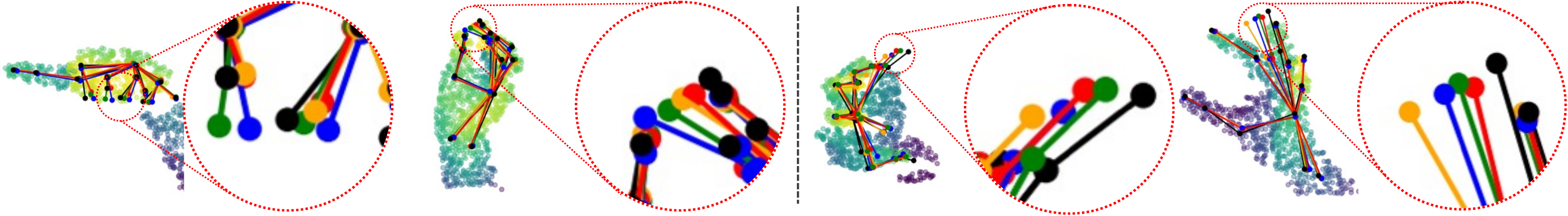}
\caption{Qualitative results of HandDAGT ablations on the NYU dataset. Ground truth is shown in black and the estimated keypoint coordinates of ablations 1) no attention, 5) attention on local points $\mathbf{Att^{local}_k}$, 6) attention on kinematic topology $\mathbf{Att^{g}_k}$ and 7) the full implementation are shown in blue, orange, green and red, respectively.}
\label{fig:abla}
\end{figure}

Table \ref{tab:abl} presents the experimental results of the ablations, which illustrate the contributions of the proposed components to the final performance. 
The comparison between ablations 1) and 2) highlights that the conventional attention mechanism alone cannot effectively process keypoint coordinate estimation and even deteriorate performance. Conversely, the proposed sigmoid attention mechanism (Equation \ref{eq.score}) leads to a drastic improvement in performance, as evidenced by the comparison between ablations 1) and 4). Furthermore, the results from ablations 3) and 4) suggest that the enhanced $\mathbf{Q}$ and $\mathbf{K}$ contribute significantly to the model's performance, as well as the conventional attention mechanism. Ablations 5) and 6) demonstrate that both $\mathbf{Att^{local}_k}$ and $\mathbf{Att^{g}_k}$ positively impact the model's performance, indicating that attention mechanisms focused on local points and kinematic topology are beneficial additions to the model architecture. We also visualize the qualitative results of the ablations in Figure \ref{fig:abla}. The qualitative results indicate that attention solely on local points can yield better performance in non-occluded cases, as illustrated in the left of Figure \ref{fig:abla}. This is attributed to the sufficient local details. Conversely, attention solely on kinematic topology demonstrates strong performance in occluded cases, as depicted in the right of Figure \ref{fig:abla}. Our full implementation surpasses all ablations by leveraging both mechanisms, thereby achieving superior performance.

\noindent
\textbf{Effectiveness of the denoising training strategy.} 
To evaluate the effectiveness of the proposed denoising training strategy, we train a model without noise disturbing during training. As reported in Table \ref{tab:denoise}, the performance is boosted with 0.08 mm of error reduction. Notably, this performance improvement does not require any additional computations, since these two models share the equivalent inference process but vary with different training strategies.

\begin{table}[t]
    \caption{Ablation experimental result. All the ablation models are trained and tested on the NYU dataset.} 
    \label{Tab}
    \begin{subtable}{.28\linewidth}
        \caption{\small Desnoising training.}
        \label{tab:denoise}
        \centering
        \begin{tabular}{c|c}
        \hline
        Denosing  & Mean keypoint\\
         training & error (mm)\\
        \hline
        $\times$ & 7.20\\
        $\surd$ & \textbf{7.12}\\
        \hline
        \end{tabular}
    \end{subtable}%
    \begin{subtable}{.4\linewidth}
        \caption{\small  Number of transformer.}
        \label{tab.stack}
        \centering
        \begin{tabular}{c|c}
        \hline
        Number of& Mean keypoint\\
        transformer & error (mm)\\
        \hline
        1 & 8.02\\
        2 & 7.15\\
        3 & \textbf{7.12}\\
        \hline
        \end{tabular}
    \end{subtable} 
    \begin{subtable}{.28\linewidth}        
        \caption{\small  Size of input patch. }
        \label{tab:patch}
        \centering
        \begin{tabular}{c|c}
        \hline
        Patch  & Mean keypoint\\
        size (pts) & error (mm)\\
        \hline
        32 & 7.20\\
        64 & \textbf{7.12}\\
        128 & 7.16\\
        \hline
        \end{tabular}
    \end{subtable} 
\end{table}

\noindent
\textbf{Number of stacked transformers.} 
Intuitively, deeper stacks of transformers have the potential to enhance performance by allowing the model to capture more intricate features. To explore this, we incrementally increased the number of stacked transformers to search for the optimal depth of the transformer stack. As outlined in Table \ref{tab.stack}, we systematically varied the depth of the stack to evaluate its impact on model performance. Based on the results, it is evident that increasing the depth of the transformer stack generally leads to performance improvements. However, the model with a 3-stack transformer exhibits only marginal enhancements. Therefore, we conclude that the optimal stack number for our model is 3, and further increasing the depth of the transformer stack does not yield significant performance gains.


\noindent
\textbf{Size of input patches.}
Indeed, the size of patches significantly influences the receptive field for the transformer, ultimately impacting the model's ability to capture detailed information. To investigate this effect, we conducted a series of ablation experiments wherein the models were trained using different patch sizes for estimation. Since the patches consist of super points, thus the size of the patch is the number of super points contained in a patch. As depicted in Table \ref{tab:patch}, the patch containing 64 super points exhibited optimal performance, whereas the experiment with patch sizes larger than 64 points showed degraded performance. This observation suggests that the receptive field of the transformer must be adequately sized to gather effective information. However, introducing a redundant patch size could lead to the inclusion of unnecessary information, thus disrupting the transformer's operation.


\section{Conclusion}
In this paper, we introduced HandDAGT, a novel adaptive graph transformer architecture designed to adaptively utilize local information and kinematic topology based on the occlusion circumstances. To improve the robustness of the proposed network, a novel denoising training strategy is introduced. Experimental results showcased that our network outperforms previous state-of-the-art methods on four challenging datasets. Moreover, extensive experiments validated the effectiveness of the components proposed in this paper. As a limitation, HandDAGT currently can not process the scenario with interacting hands and requires redundant computation due to the 2D feature extraction. Addressing these limitations could involve extending our research to include bidirectional learning and developing a lightweight 2D model. These avenues remain open for future research and represent potential areas for enhancing the capabilities and efficiency of HandDAGT.

\section*{Acknowledgement} This work was supported by the Ministry of Science and ICT (MSIT) of Korea under the National Research Foundation (NRF) grant (2022R1A4A3032913) and the Institute of Information and Communication Technology Planning Evaluation (IITP) grants (IITP-2019-0-00421, IITP-2023-2020-0-01821, IITP-2021-0-02052, IITP-2021-0-02068), and by the Technology Innovation Program (RS-2023-00235718) funded by the Ministry of Trade, Industry \& Energy (1415187474). Wencan Cheng was partly supported by the China Scholarship Council (CSC).
%
%

\end{document}